\documentclass{article}

\PassOptionsToPackage{numbers}{natbib}
\usepackage{ml_institute}

\usepackage{amsmath,amsfonts,bm}

\def\eqref#1{equation~\ref{#1}}

\def\1{\bm{1}}

\def\vb{{\bm{b}}}

\def\vh{{\bm{h}}}

\def\vk{{\bm{k}}}

\def\vm{{\bm{m}}}
\def\vn{{\bm{n}}}
\def\vo{{\bm{o}}}

\def\vq{{\bm{q}}}

\def\vv{{\bm{v}}}
\def\vw{{\bm{w}}}
\def\vx{{\bm{x}}}

\def\mC{{\bm{C}}}

\def\mK{{\bm{K}}}

\def\mQ{{\bm{Q}}}

\def\mW{{\bm{W}}}
\def\mX{{\bm{X}}}
\def\mY{{\bm{Y}}}

\DeclareMathAlphabet{\mathsfit}{\encodingdefault}{\sfdefault}{m}{sl}
\SetMathAlphabet{\mathsfit}{bold}{\encodingdefault}{\sfdefault}{bx}{n}
\newcommand{\tens}[1]{\bm{\mathsfit{#1}}}

\def\tX{{\tens{X}}}

\usepackage{booktabs}
\usepackage{graphicx}
\usepackage[table]{xcolor}
\usepackage{multirow}
\usepackage{xcolor}
\usepackage{subcaption}
\usepackage{colortbl}
\usepackage{url}
\definecolor{cbred}{rgb}{0.9255, 0.6157, 0.5922}
\definecolor{cbgreen}{rgb}{0.2275, 0.6863, 0.4275}

\definecolor{deemph}{gray}{0.6}

\usepackage{pifont}
\newcommand{\xmark}{\ding{55}}
\newcommand{\cmark}{\ding{51}}

\usepackage{fancyhdr}

\definecolor{defaultcolor}{rgb}{0.8666, 0.8666, 0.8666}
\newcommand{\default}[1]{\cellcolor{defaultcolor}{#1}}

\definecolor{cbgray}{rgb}{0.86666667, 0.86666667, 0.86666667}
\definecolor{cbpink}{rgb}{0.97647059, 0.6627451, 0.7372549}
\definecolor{cbyellow}{rgb}{0.92941176, 0.87058824, 0.5333}
\definecolor{cblightblue}{rgb}{0.6, 0.87058824, 1.}
\definecolor{cborange}{rgb}{0.93333333, 0.53333333, 0.4}
\definecolor{cbblue}{rgb}{0.46666667, 0.6627451, 0.86666667}
\colorlet{cborange05}{cborange!50!white}
\colorlet{cbblue05}{cbblue!50!white}
\colorlet{cbgray05}{cbgray!50!white}
\colorlet{cbpink05}{cbpink!50!white}
\colorlet{cbyellow05}{cbyellow!50!white}
\colorlet{cblightblue05}{cblightblue!50!white}

\usepackage{float}

\title{Vision-LSTM: xLSTM as Generic Vision Backbone}

\author{
Benedikt Alkin$~^{1,2}$ \quad
Maximilian Beck$~^{1,3}$ \quad
Korbinian Pöppel$~^{1,3}$ \quad \\ \bf
Sepp Hochreiter$~^{1,2,3}$ \quad
Johannes Brandstetter$~^{1,2}$\\ \\
$~^{1}$~ELLIS Unit Linz, Institute for Machine Learning, JKU Linz, Austria\\
$~^{2}$~Emmi AI GmbH, Linz, Austria\\
$~^{3}$~NXAI GmbH, Linz, Austria\\
\texttt{\{alkin,brandstetter\}@ml.jku.at}
}

\begin{document}

\maketitle

\begin{abstract}%

Transformers are widely used as generic backbones in computer vision, despite initially introduced for natural language processing. Recently, the Long Short-Term Memory (LSTM) has been extended to a scalable and performant architecture -- the xLSTM -- which overcomes long-standing LSTM limitations via exponential gating and parallelizable matrix memory structure. In this paper, we introduce Vision-LSTM (ViL), an adaption of the xLSTM building blocks to computer vision. ViL comprises a stack of xLSTM blocks where odd blocks process the sequence of patch tokens from top to bottom while even blocks go from bottom to top.
ViL achieves strong performances on classification, transfer learning and segmentation tasks as well as a beneficial pre-training cost-to-performance trade-off.
Experiments show that ViL holds promise to be further deployed as new generic backbone for computer vision architectures. \\
Project page: \url{https://nx-ai.github.io/vision-lstm/}

\end{abstract}

\begin{figure}[h]
\centering
\includegraphics[width=\linewidth]{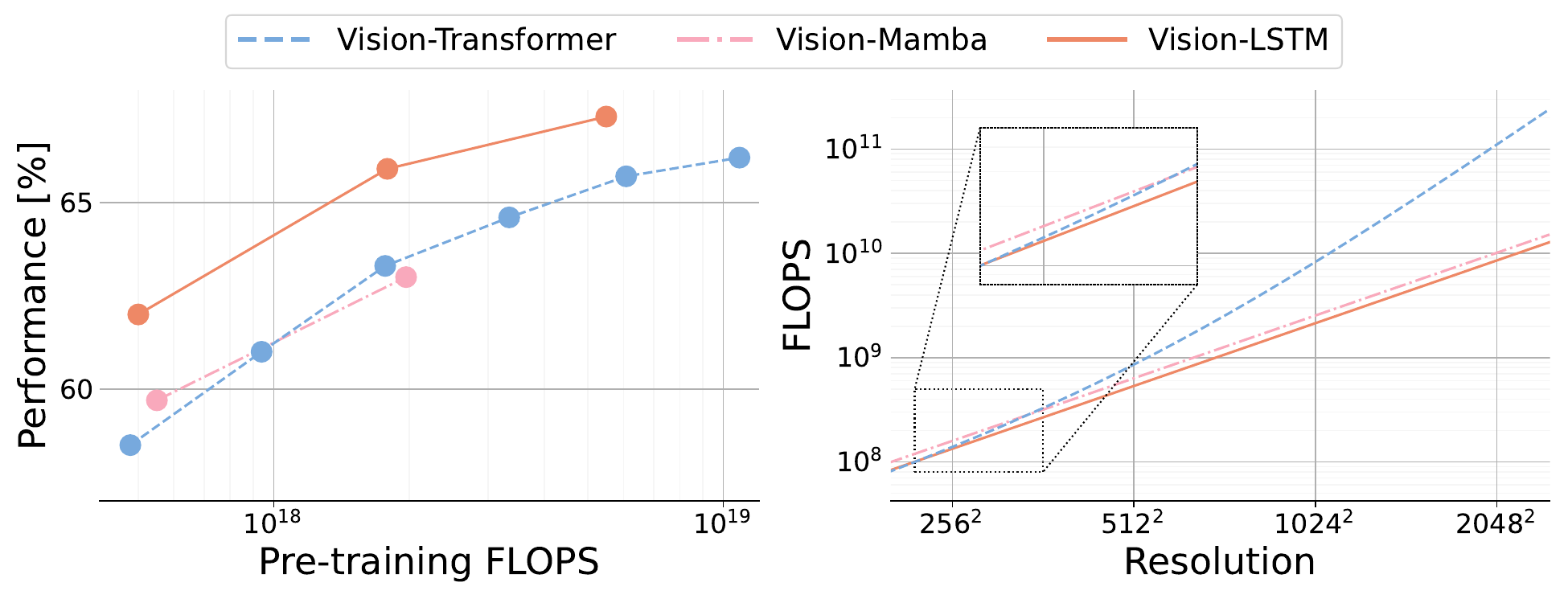}
\caption{ The efficient and scalable design of Vision-LSTM shows strong performances, uses less FLOPS than Transformer/Mamba counterparts and scales linear to higher resolutions. Performance is averaged over ImageNet accuracy, ADE20K mIoU and VTAB-1K accuracy.}
\label{fig:ViL}
\end{figure}

\let\thefootnote\relax\footnotetext{Published as a conference paper at ICLR 2025}

\section{Introduction}

Language modeling architectures --- such as Transformers~\citep{vaswani2017attention,achiam2023gpt,team2023gemini} or more recently State Space Models~\citep{Gu:21,Gupta:22} such as Mamba~\citep{gu2023mamba} --- are commonly adapted to the domain of computer vision to make use of their powerful modeling capabilities. However, in natural language processing, an input sentence is typically encoded into tokens that represent words or common subwords~\citep{bostrom2020byte} via a discrete vocabulary. To encode images into a set of tokens, Vision Transformer~\citep{dosovitskiy2021vit} (ViT) proposed to group an input image into non-overlapping patches (of e.g.\ 16x16 pixel), linearly project them into a sequence of so-called patch tokens and add positional information to these tokens. This sequence can then be processed by language modeling architectures.

The Extended Long Short-Term Memory (xLSTM) family~\citep{beck2024xlstm} was recently introduced as a new architecture for language modeling. It demonstrates the resurgence of LSTM in the LLM era, performing favorably against the likes of Transformers and State Space Models (SSMs). Analogous to existing vision versions of Transformers or State Space Models, e.g.,\ ViT~\citep{dosovitskiy2021vit} or Vision Mamba~\citep{zhou2024visionmamba}, which have produced great results in various computer vision tasks~\citep{singh2023maewsp,kirillov2023segmentanything,oquab2023dinov2,peebles2023dit,alkin2024mimrefiner}, we introduce Vision LSTM (ViL) -- a generic computer vision backbone that uses xLSTM blocks as its core components.
To adjust xLSTM (an autoregressive model) to computer vision (an often non-autoregressive domain), we employ a stack of alternating mLSTM blocks~\citep{beck2024xlstm} where odd blocks process patches row-wise from top left to bottom right and even blocks go from bottom right to top left. This simple alternating design allows ViL to efficiently process non-sequential inputs, such as images, without introducing additional computations.

\begin{figure}[t!]
\centering
\includegraphics[width=\linewidth]{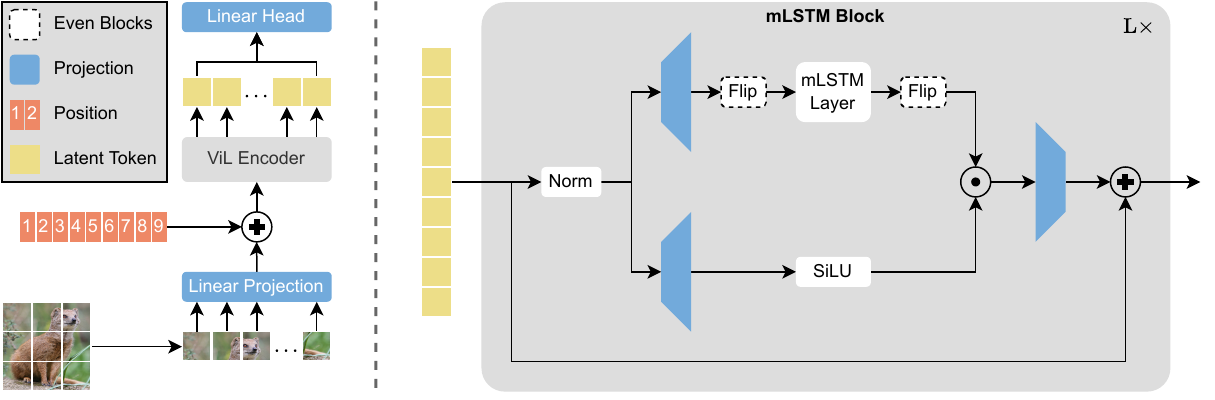}
\caption{ Schematic overview of Vision-LSTM (ViL). Following ViT~\citep{dosovitskiy2021vit}, an input image is split into patches and linearly projected. Then, a learnable vector is added per position to the patches, producing a sequence of patch tokens. This sequence is then processed by alternating mLSTM blocks where even blocks flip the sequence before and after the mLSTM layer. For classification, ViL uses the concatenation of the first and the last patch as input to a linear classification head. ViL is an isotropic architecture, i.e., all blocks have the same input and output dimension and no downsampling layers are used except the initial patch embedding. Projection layers process each patch individually and the mLSTM exchanges information between patches. }
\label{fig:ViL}
\end{figure}

Similar to vision adaptions of SSMs~\citep{liu2024vmamba,zhou2024visionmamba,wang2024visionmambar}, ViL can exhibit linear computational and memory complexity w.r.t.\ sequence length which makes it appealing for tasks that benefit from high-resolution images such as medical imaging~\citep{chen2021transunet,hatamizadeh2022unetr,valanarasu2021medicaltransformer,xu2024wholeslidefoundationmodel}, segmentation~\citep{kirillov2023segmentanything,cheng2022mask2former}, or physics simulations~\citep{bi2023accurate,nguyen2023climax,bodnar2024aurora,alkin2024upt}. In contrast, ViT's computational complexity scales quadratically due to the self-attention mechanism, rendering them costly to apply to high-resolution tasks.

Our contributions summarize as follows:
\begin{itemize}
\item We introduce Vision-LSTM (ViL), an adaption of the mLSTM to computer vision tasks that can serve as a generic vision backbone with linear complexity.
\item We show modeling capacity and generalization in the common vision benchmark of pre-training models on ImageNet-1K, followed by fine-tuning on transfer classification and semantic segmentation tasks.
\item We ablate various architectural design choices to evaluate their impact on performance and provide insights into the model design.
\item We discuss potential future directions and current limitations that, once addressed, will improve ViL even further.
\end{itemize}

\section{Method}

Vision-LSTM (ViL) introduces xLSTM~\citep{beck2024xlstm} to computer vision, similar to other vision adaptions of sequence modeling architectures, e.g., Vision Transformers~\citep{dosovitskiy2021vit}, Vision Mamba~\citep{zhou2024visionmamba}, or Vision RWKV~\citep{duan2024vrwkv}.

\subsection{Preliminaries}

In the notation of sequence modeling, we consider a series of input vectors $\vx_t \in \mathbb{R}^D$. This series is created by reshaping an image $\tilde{\tX} \in \mathbb{R}^{H_I\times W_I \times C_\text{in}}$ into a sequence of flattened 2D patches $\bar{\mX} \in \mathbb{R}^{T \times (H_P \cdot W_P \cdot C_\text{in})}$ and then projected to $\mX \in \mathbb{R}^{T\times D}$ via a shared linear projection. $D$ is the hidden dimension, $(H_I, W_I)$ is the image resolution, $C_\text{in}$ is the number of image channels, $T$ is the number of patches and $(H_P, W_P)$ is the patch size. After creating a sequence of patches, ViL iteratively refines the features of the patch sequence by processing it with a stack of mLSTM blocks where the sequence is flipped within every second block.

The key innovations of the mLSTM~\citep{beck2024xlstm} are the enhanced storage capacity compared to the classical LSTM~\citep{hochreiter1997long} by using a matrix memory cell $\mC \in \mathbb{R}^{d \times d}$ instead of a scalar memory cell $c \in \mathbb{R}$ and introducing exponential gates (instead of sigmoid gates) to the input and forget gates, where $d$ is the hidden dimension within the mLSTM block (typically $d = 2D$).

Intuitively, the mLSTM is a more expressive and faster version of the classical LSTM that can be efficiently parallelized on modern hardware. In ViL, the mLSTM is used to process dependencies between patches, similar to how the attention exchanges information between patches in a ViT. The mLSTM is embedded into a gated MLP architecture, as shown on the right of Figure~\ref{fig:ViL}, where the weight matrices of the MLP process each patch individually and the mLSTM exchanges information between patches. For completeness, we outline the forward pass of the mLSTM in the following paragraphs.

The mLSTM~\citep{beck2024xlstm} is a recurrent neural network, which maps a state $(\vh_{t-1}, \mC_{t-1},\vn_{t-1})$ to a successor state $(\vh_{t}, \mC_{t},\vn_{t})$ given input $\vx_{t-1}$. Thereby, $\vh_t \in \mathbb{R}^d$ denotes the hidden state, $\mC_t \in \mathbb{R}^{d \times d}$ is the cell state and $\vn_t \in \mathbb{R}^d$ corresponds to a normalizer state.
The full forward pass of the mLSTM is as follows~\citep{beck2024xlstm}:

\begin{align}
\mC_t \ &= \  f_t \ \mC_{t-1} \ + \  i_t \ \vv_t \ \vk_t^\top &&&\text{cell state} \\
\vn_t \ &= \  f_t \ \vn_{t-1} \ + \  i_t \ \vk_t &&&\text{normalizer state} \\
\vh_t  \ &= \ \vo_t \ \odot \ \tilde{\vh}_t \  &\tilde{\vh}_t \ &= \   \mC_t \vq_t \ / \  \max \left\{ |{\vn_t^\top \vq_t}|, 1 \right\}  &\text{hidden state} \\
\vq_t \ &= \ \mW_q \ \vx_t \ + \ \vb_q  &&&\text{query input} \\
\vk_t \ &= \ \frac{1}{\sqrt{d}} \mW_k \ \vx_t \ + \ \vb_k  &&&\text{key input} \\
\vv_t \ &= \ \mW_v \ \vx_t \ + \ \vb_v  &&&\text{value input} \\
i_t \   &= \ \exp  \! \big( \tilde{i}_t \big) \  &\tilde{i}_t    \ &= \ \vw^\top_{i} \ \vx_t \ + \  b_{i} &\text{input gate} \\
f_t \   &= \ \exp  \! \big( \tilde{f}_t \big) \  &\tilde{f}_t    \ &= \ \vw^\top_{f} \ \vx_t  \ + \  b_{f}   & \text{forget gate} \\
\vo_t \  &= \ \sigma \big( \tilde{fo}_t \big)     \  &\tilde{\vo}_t  \ &= \ \mW_{\vo} \ \vx_t \ + \ \vb_{\vo} &\text{output gate}
\end{align}

As exponential activation functions can lead to large activations, the input and forget gates are stabilized with an additional state $m_t$:

\begin{align}
m_t &= \max  \! \Big( \log(f_t) + \vm_{t-1}, \log(f_t) \Big) & \phantom{a + b + c} && \text{stabilizer state} \\
i'_t &= \exp \! \Big( \log(i_t) - m_t \Big) = \exp \Big(\tilde{i} - m_t \Big) &&& \text{stabilized input gate} \\
f'_t &= \exp \! \Big( \log(f_t) + m_{t-1} - m_t \Big) &&& \text{stabilized forget gate}
\end{align}

As the mLSTM has no memory mixing, i.e, interactions between hidden states from one timestep to the next, it can be fully parallelized for fast computation on modern hardware.
For a detailed discussion and theory of the cell state update, further details to the mLSTM we refer to the original work \citep{beck2024xlstm}.

\subsection{Vision-LSTM (ViL)} \label{sec:method_vil}

Vision-LSTM (ViL) is a generic backbone for computer vision tasks, which is residually built from mLSTM blocks, as visualized in Figure~\ref{fig:ViL}. Following ViT~\citep{dosovitskiy2021vit}, ViL first splits an image into non-overlapping patches via a shared linear projection, then adds learnable positional embeddings to each patch token. At the core of ViL are alternating mLSTM blocks, which are fully parallelizable and equipped with a matrix memory combined with a covariance update rule. Odd mLSTM blocks process patch tokens from top left to bottom right while even blocks go from bottom right to top left.

Formally, the forward pass of a pair of ViL blocks is:

\begin{align}
\mY' &= \mX + \text{Block}_\theta(\mX) \\
\mY &= \mY' + \text{Flip}(\text{Block}_\phi(\text{Flip}(\mY')))
\end{align}

Where ``Flip'' reverses the sequence and ``$\text{Block}_\theta$'' and ``$\text{Block}_\phi$'' corresponds to mLSTM blocks with parameters $\theta$ and $\phi$ (shown in Figure~\ref{fig:ViL}, right).

A key motivation of ViL is that the autoregressive mLSTM can operate in a recurrent, parallel or chunkwise mode, each with distinct FLOPS and runtime characteristics. Given a sequence length $T$ and hidden dimension $d$,
the complexity of the recurrent mode is $\mathcal{O}(Td^2)$ and needs to be processed sequentially, whereas the parallel mode has complexity $\mathcal{O}(T^2d)$ and is fully parallelizable. The chunkwise mode combines the advantages of the other modes by introducing a chunksize $S$ where the parallel mode is used within chunks and the recurrent mode between chunks. This allows high parallelization, minimal operations and linear scaling with $T$. Complexity wise, the chunkwise mode has $\mathcal{O}(\frac{T}{S}S^2d + \frac{T}{S}d^2)$ or  $\mathcal{O}(TSd + \frac{T}{S}d^2)$ where $\frac{T}{S}$ corresponds to the number of chunks.

\section{Experiments} \label{sec:experiments}

\begin{figure}[h]
\centering
\hfill
\begin{minipage}{0.32\linewidth}
\centering
\includegraphics[width=\linewidth]{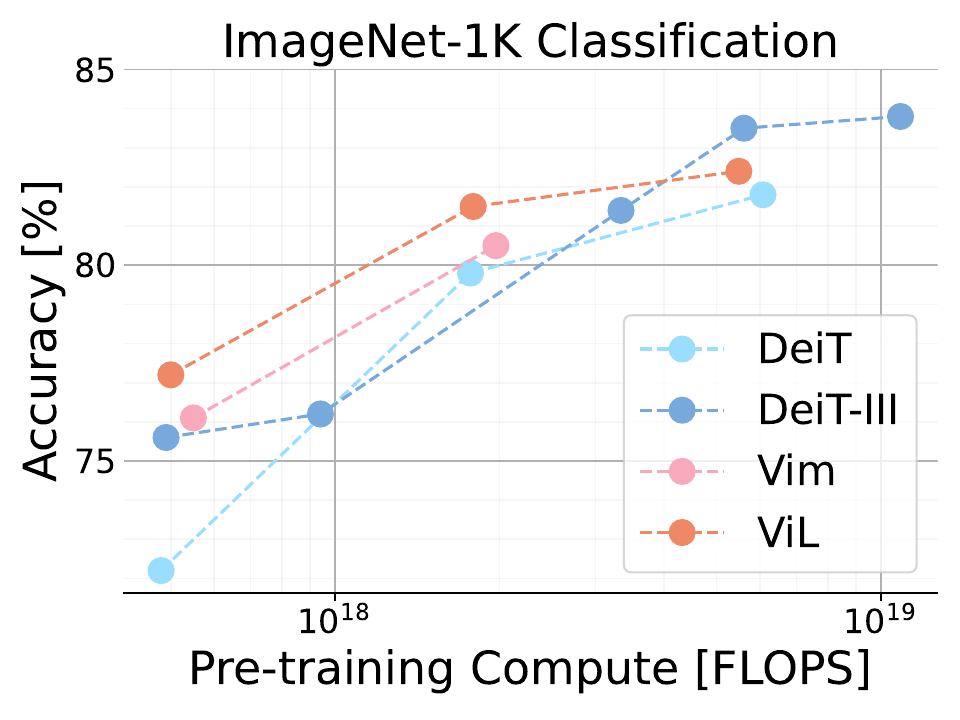}
\end{minipage}
\hfill
\begin{minipage}{0.32\linewidth}
\centering
\includegraphics[width=\linewidth]{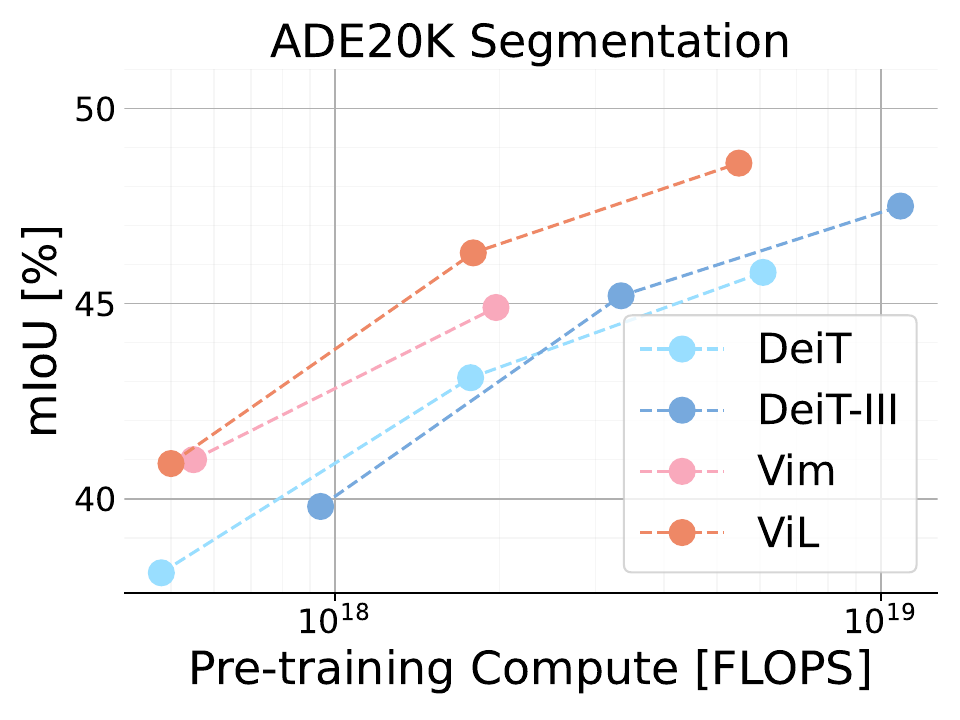}
\end{minipage}
\hfill
\begin{minipage}{0.32\linewidth}
\centering
\includegraphics[width=\linewidth]{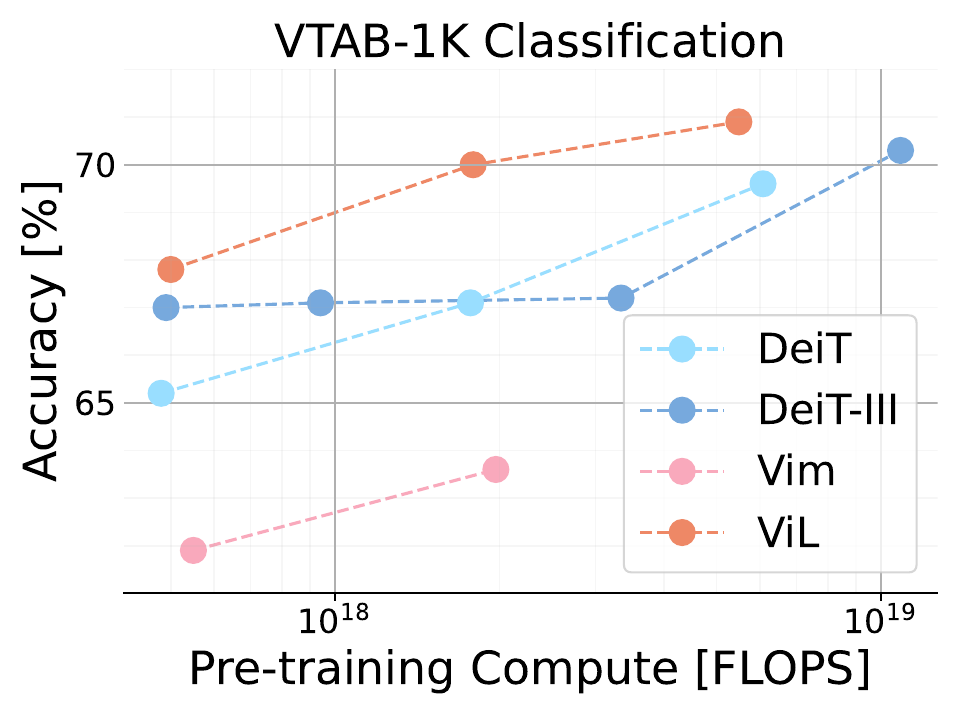}
\end{minipage}
\hfill
\hspace{0.01cm}
\caption{ Performance overview of ImageNet-1K pre-trained models in relation to pre-training compute. ViL shows strong performances across classification (ImageNet-1K), semantic segmentation (ADE20K) and transfer classification (VTAB-1K) tasks. }
\label{fig:flops_over_performance}
\end{figure}

We pre-train models on ImageNet-1K~\citep{deng2009imagenet}, which contains 1.3M training images and 50K validation images where each image belongs to one of 1000 classes.  ViL models are trained for 800 epochs (tiny) or 400 epochs (small, base) on 192x192 resolution with a learning rate of 1e-3 using a cosine decay schedule. Afterwards, the model is fine-tuned on 224x224 resolution for 20 epochs using a learning rate of 1e-5. Detailed hyperparameters can be found in Appendix Table~\ref{tab:hyperparams_ViL_pretrain}.

We then transfer the pre-trained models to serveral benchmark tasks: ImageNet-1K classification on the validation set, ADE20K~\citep{zhou2019ade20k} semantic segmentation and VTAB-1K~\citep{zhai2019vtab} classification. These benchmarks evaluate global image understanding (ImageNet-1K), semantic local and global understanding (ADE20K) and few-shot generalization to a diverse set of 19 VTAB-1K classification datasets, which include natural images, specialized imagery (medical and satellite) and structured tasks (camera angle prediction, depth estimation, object counting, \dots).

Figure~\ref{fig:flops_over_performance} shows an overview of performance metrics in relation to total pre-training compute where ViL performs favorably against heavily optimized transformer protocols (DeiT, DeiT-III) and Vision Mamba (Vim). Detailed results are presented in the following sections.

As ViTs are well established in the vision community, they underwent multiple optimization cycles over the years~\citep{dosovitskiy2021vit,touvron2021deit,touvron2022threethingsaboutvit,touvron2021cait,touvron2022deit3}. Therefore, a vast part of the hyperparameter space for pre-training ViTs has been explored. Since this work is the first to apply xLSTM to computer vision, considerably less effort has been put into hyperparameter tuning and architecture optimization, suggesting that future work could improve ViL even further.

\subsection{ImageNet-1K Classification}

Table~\ref{tab:imagenet1k} relates parameter counts and FLOPS to validation accuracy after pre-training on ImageNet-1K. ViL outperforms heavily optimized ViT protocols and other backbones on the tiny and small scale. While ViL does not outperform all other models on the base scale, evaluations on downstream tasks (as shown later in Table~\ref{tab:ade20k} and Table~\ref{tab:vtab1k}) show that ViL-B still learns strong features, particularly for semantic segmentation and structured tasks.

\begin{table}[t]
\centering
\small
\begin{tabular}{lcccc}
Model & Epochs & \#Params & FLOPS & IN-1K \\
\hline
DeiT-T~\citep{touvron2021deit} & 300 & 6M & 1.3G & 72.2\\
DeiT-II-T~\citep{touvron2022threethingsaboutvit} & 400 & 6M & 1.3G & 73.5 \\
DeiT-III-T (reimpl.) & 800+20 & 6M & 1.3G & 76.2 \\
VRWKV-T~\citep{duan2024vrwkv} & 300 & 6M & 1.2G & 75.1 \\
Vim-T~\citep{zhou2024visionmamba} & 300 & 7M & 1.5G & 76.1 \\
Mamba\textsuperscript{\textregistered}-T~\citep{wang2024visionmambar} & 280+20 & 9M & 1.6G & 77.4 \\
\default{ViL-T} & \default{800+20} & \default{6M} & \default{1.3G} & \default{\textbf{78.3}} \\
\hline
DeiT-S~\citep{touvron2021deit} & 300 & 22M & 4.6G & 79.8 \\
DeiT-II-S~\citep{touvron2022threethingsaboutvit} & 400 & 22M & 4.6G & 80.7 \\
DeiT-III-S~\citep{touvron2022deit3} & 800+20 & 22M & 4.6G & 81.4 \\
ConvNeXt-S (\textit{iso.})~\citep{liu2022convnext} & 300 & 22M & 4.3G & 79.7 \\
VRWKV-S~\citep{duan2024vrwkv} & 300 & 24M & 4.6G & 80.1 \\
Vim-S~\citep{zhou2024visionmamba} & 300 & 26M & 5.3G & 80.5 \\
Mamba\textsuperscript{\textregistered}-S~\citep{wang2024visionmambar} & 280+20 & 28M & 5.5G & 81.1 \\
\default{ViL-S} & \default{400+20} & \default{23M} & \default{4.7G} & \default{\textbf{81.5}} \\
\hline
DeiT-B~\citep{touvron2021deit} & 300 & 86M & 17.6G & 81.8 \\
DeiT-II-B~\citep{touvron2022threethingsaboutvit} & 400 & 86M & 17.6G & 82.7 \\
DeiT-III-B~\citep{touvron2022deit3} & 800+20 & 86M & 17.6G & \textbf{83.7} \\
ConvNeXt-B (\textit{iso.})~\citep{liu2022convnext} & 300 & 87M & 16.9G & 82.0 \\
VRWKV-B~\citep{duan2024vrwkv} & 300 & 94M & 18.2G & 82.0 \\
Mamba\textsuperscript{\textregistered}-B~\citep{wang2024visionmambar} & 280+20 & 99M & 20.6G & 82.9 \\
\default{ViL-B} & \default{400+5} & \default{89M} & \default{17.9G} & \default{82.4} \\
\end{tabular}
\vspace{1em}
\caption{ImageNet-1K pre-training accuracy. All models use a patch size of 16x16 with 224x224 resolution at most. Models with ``+'' in their ``Epochs'' column pre-train on lower resolution followed by fine-tuning on 224x224 resolution for some epochs. ViL performs favorably against an isotropic convolutional architecture (ConvNeXt) and vision adaptions of transformers (DeiT series), RWKV (VRWKV) and Mamba (Vim, Mamba\textsuperscript{\textregistered}). Appendix Table~\ref{tab:robustness} confirms these results on OOD and robustness evaluations of these classifiers.
}
\label{tab:imagenet1k}
\end{table}

\subsection{ADE20K Semantic Segmentation}

Table~\ref{tab:ade20k} shows results for transferring ImageNet-1K pre-trained models to ADE20K~\citep{zhou2019ade20k} semantic segmentation using UperNet~\citep{xiao2018upernet}.
Also here, ViL shows strong performances across the board, even outperforming DeiT-III-B despite the lower ImageNet-1K accuracy of ViL-B. The high resolution of the ADE20K segmentation task (512x512) results in a total of 1024 patch tokens where the quadratic complexity of self-attention is significantly more expensive than the linear complexity of the mLSTM, resulting in much fewer FLOPS for ViL. Additionally, the efficient alternating block design results in lower FLOPS than Mamba-based vision models (which also have linear complexity).

\begin{table}[h]
\centering
\small
\begin{tabular}{lcccccc}
& &  & \multicolumn{2}{c}{\textbf{Single-scale}} & \multicolumn{2}{c}{\textbf{Multi-scale}} \\
\cmidrule(rl){4-5}\cmidrule(rl){6-7}
Model & \#Params & FLOPS & mIoU & ACC & mIoU & ACC \\
\hline
DeiT-T & 10M & 10.4G & 38.1 & 78.2 & 40.3 & 79.9 \\
DeiT-III-T & 10M & 10.4G & 39.8 & 79.2 & 42.2 & 80.7 \\
Vim-T & 13M & 7.7G & 41.0 & - & - & - \\
\default{ViL-T} & \default{11M} & \default{\textbf{6.6G}} & \default{\textbf{41.2}} & \default{\textbf{80.2}} & \default{\textbf{43.1}} & \default{\textbf{81.3}} \\
\hline
DeiT-S & 41M & 31.7G & 43.1 & 80.7 & 45.2 & 81.8 \\
DeiT-III-S & 41M & 31.7G & 45.2 & 81.5 & 46.3 & 82.3 \\
Vim-S & 46M & 27.3G & 44.9 & - & - & - \\
Mamba\textsuperscript{\textregistered}-S & 56M & 27.6G & 45.3 & - & - & - \\
\default{ViL-S} & \default{42M} & \default{\textbf{24.4G}} & \default{\textbf{46.3}} & \default{\textbf{82.0}} & \default{\textbf{47.9}} & \default{\textbf{82.9}} \\
\hline
DeiT-B & 113M & 107.0G & 45.8 & 82.1 & 47.0 & 82.9 \\
DeiT-III-B & 113M & 107.0G & 47.5 & 82.6 & 49.0 & 83.3 \\
Mamba\textsuperscript{\textregistered}-B & 132M & 102.8G & 47.7 & - & - & - \\
\default{ViL-B} & \default{115M} & \default{\textbf{93.6G}} & \default{\textbf{48.6}} & \default{\textbf{82.8}} & \default{\textbf{49.6}} & \default{\textbf{83.3}} \\
\end{tabular}
\vspace{1em}
\caption{
Semantic segmentation results on ADE20K~\citep{zhou2019ade20k} using UperNet~\citep{xiao2018upernet}. We report mean intersection over union (mIoU) and pixelwise accuracy (ACC) for single- and multi-scale evaluation.
Models are trained for 160K updates with a batchsize of 16 on 512x512 resolution. We use a feature pyramid consisting of rescaled feature maps after the 4th, 6th, 8th and final block. 
Detailed hyperparameters are listed in Appendix Table~\ref{tab:hyperparams_finetune_ade20k}. FLOPS are calculated only from the backbone at 512x512 resolution as all models use the same segmentation head.
}
\label{tab:ade20k}
\end{table}

\subsection{VTAB-1K Transfer Classification}

\begin{table}[h]
\centering
\small
\begin{tabular}{lcccccc}
Model & \#Params & FLOPS & Natural & Specialized & Structured & Average \\
\hline
DeiT-T & 6M & 1.3G & 69.2 & 82.0 & 53.3 & 65.2 \\
DeiT-III-T & 6M & 1.3G & 71.9 & 82.6 & 55.2 & 67.1 \\
Vim-T & 7M & 1.5G & 68.0 & 80.7 & 47.1 & 61.9 \\
\default{ViL-T} & \default{6M} & \default{1.3G} & \default{\textbf{73.6}} & \default{\textbf{83.4}} & \default{\textbf{56.1}} & \default{\textbf{68.3}} \\
\hline
DeiT-S & 22M & 4.6G & 73.3 & 83.8 & 53.2 & 67.1 \\
DeiT-III-S & 22M & 4.6G & 75.0 & 83.2 & 52.3 & 67.2 \\
Vim-S & 26M & 5.3G & 69.6 & 81.7 & 49.4 & 63.6 \\
\default{ViL-S} & \default{23M} & \default{4.7G} & \default{\textbf{75.3}} & \default{\textbf{84.3}} & \default{\textbf{58.3}} & \default{\textbf{70.0}} \\
\hline
DeiT-B & 86M & 17.6G & 76.5 & \textbf{85.2} & 55.7 & 69.6 \\
DeiT-III-B & 86M & 17.6G & \textbf{77.6} & 84.8 & 56.6 & 70.3 \\
\default{ViL-B} & \default{89M} & \default{17.9G} & \default{76.6} & \default{84.7} & \default{\textbf{59.1}} & \default{\textbf{70.9}} \\
\end{tabular}
\vspace{1em}
\caption{Transfer classification accuracies on the VTAB-1K~\citep{zhai2019vtab} benchmark using ImageNet-1K pre-trained models. VTAB-1K consists of 19 datasets split into 7 natural, 4 specialized and 8 structured datasets. We show averages per category and the average accuracy over all 19 datasets (Appendix Table~\ref{tab:vtab1k_full} lists all individual accuracies). ViL shows strong generalization performance, outperforming heavily optimized ViT protocols and Vim on the full VTAB-1K benchmark. ViL performs exceptionally well on the structured category. We tune the learning rate for each model and dataset on the validation set and report the average testset accuracy over 5 seeds. Appendix Table~\ref{tab:hyperparams_finetune_vtab1k} lists further hyperparameters.  }
\label{tab:vtab1k}
\end{table}

Table~\ref{tab:vtab1k} shows transfer classification results for ImageNet-1K pre-trained models on the VTAB-1K~\citep{zhai2019vtab} benchmark. VTAB-1K consists of 19 datasets split into 7 natural datasets (such as CIFAR100~\citep{krizhevsky2009cifar} or Caltech101~\citep{feifei2006caltech101}), 4 specialized datasets (medical imaging~\citep{veeling2018patchchamelyon,kaggle2015diabeticretinopathy} and remote sensing~\citep{helber2019eurosat,cheng2017resisc45}) and 8 structured datasets (with tasks such as object counting~\citep{johnson2017clevr} or binned depth estimation~\citep{geiger2013kitti}). We follow common practices and tune the learning rate per model and dataset on the validation set followed by training with the best learning rate on the union of train and validation set. The performance metric is the average testset accuracy over 5 seeds. ViL shows strong transfer classification performance outperforming all other models on the average over all 19 datasets. ViL performs particularly well on the structured datasets where ViL-B outperforms DeiT-III-B despite ViL-B having lower ImageNet-1K accuracy.

\section{Ablation Studies}

We ablate various design choices of ViL by training ViL-T models for 100 epochs on ImageNet-1K in 224x224 resolution, other hyperparameters follow the ones from Section~\ref{sec:experiments} (see also Appendix~\ref{sec:vil_hyperparameters}). We then report the validation accuracy on ImageNet-1K and fine-tune the model on ADE20K to ensure that design choices are not overfitted to classification. We also use a reduced segmentation pipeline where we use a linear segmentation head and train for 40K updates using a batch size of 16 (other hyperparameters follow Appendix~\ref{tab:hyperparams_finetune_ade20k}).

\subsection{Architectural Design}

We consider various architecture design choices in Table~\ref{tab:ablations}.

\begin{table}[h]
\centering
\subfloat[
\textbf{Traversal Directions}
]{
\centering
\begin{minipage}{0.45\linewidth}
\begin{center}
\begin{tabular}{lcc}
Directions & IN1K & ADE20K \\
\hline
Uni-dir. & 72.2 & 28.6  \\
\default{Bi-dir.} & \default{73.7} & \default{31.7} \\
Quad-dir. & \textbf{73.8} & \textbf{33.1}  \\
Oct-dir. & 73.5 & 32.4 \\
\\
\end{tabular}
\end{center}
\end{minipage}
}
\subfloat[
\textbf{QK Convolution}
]{
\centering
\begin{minipage}{0.45\linewidth}
\begin{center}
\begin{tabular}{lcc}
Convolution & IN1K & ADE20K \\
\hline
None & 72.3 & 29.2 \\
Causal-Conv1D & 72.8 & 27.8 \\
Conv1D & 72.8 & 28.4 \\
\default{Conv2D} & \default{\textbf{73.7}} & \default{\textbf{31.7}} \\
\\
\end{tabular}
\end{center}
\end{minipage}
}
\\
\subfloat[
\textbf{Positional Embedding}
]{
\centering
\begin{minipage}{0.45\linewidth}
\begin{center}
\begin{tabular}{ccc}
Pos. Embed. & IN1K & ADE20K \\
\hline
\xmark & \textbf{73.7} & 31.0 \\
\default{\cmark} & \default{\textbf{73.7}} & \default{\textbf{31.7}} \\
\end{tabular}
\end{center}
\end{minipage}
}
\subfloat[
\textbf{Concurrency}
]{
\centering
\begin{minipage}{0.45\linewidth}
\begin{center}
\begin{tabular}{lcc}
Concurrency & IN1K & ADE20K \\
\hline
\default{Sequential} & \default{\textbf{73.7}} & \default{\textbf{31.7}} \\
Parallel & 73.0 & 30.6 \\
\end{tabular}
\end{center}
\end{minipage}
}
\caption{ Architecture design ablation studies.  \colorbox{defaultcolor}{Default settings}}
\label{tab:ablations}
\end{table}

\begin{figure}[H]
\centering
\includegraphics[width=0.7\linewidth]{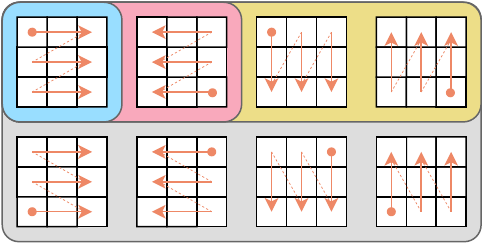}
\caption{ \colorbox{cblightblue05}{Uni-directional}, \colorbox{cbpink05}{bi-directional}, \colorbox{cbyellow05}{quad-directional} and \colorbox{cbgray05}{oct-directional} traversal paths. Squares represent individual patch tokens. Traversal starts at the circle and goes in direction of the arrow, if no further patches are in a row/column, the traversal continues in the next row/column as indicated by the dashed line. }
\label{fig:directions}
\end{figure}

\paragraph{(a) Traversal Directions} Traversing the sequence in at least two directions greatly improves performance due to the non-causal 2D structure of images. Adding column-wise traversal directions (Quad-dir.) could even further improve semantic segmentation performance. Additionally using 4 instead of 2 starting positions (Oct-dir.) shows no benefit. Note that all variants have the same amount of FLOPS due to sequential application of different directions. Directions are visualized in Figure~\ref{fig:directions}.

We use ``Bi-dir.'' for our final models due to current technical limitations which would slow down training on more than 2 directions. This limitation comes from the current lack of optimized hardware implementations of the mLSTM (e.g., CUDA kernels) where we instead rely on \texttt{torch.compile}, a generic speed optimization method from {PyTorch}~\citep{paszke2019pytorch}, to optimize computations. Our implementation of quad- and oct-directional traversals is not compatible with \texttt{torch.compile}, which results in approximately double the runtime. We therefore train all models from Section~\ref{sec:experiments} with ``Bi-dir.''. Note that this is merely a technical limitation, not a methodical one and the ablation study suggest that future ViL models could be even better using a quad-directional design.

\paragraph{(b) QK Convolution} The mLSTM block design uses a causal 1D convolution to aggregate local context to improve storage/retrieval to/from the cell state $\mC$.
This is done by applying a convolution layer to $\mX$ before projecting it to $\mQ$ with $\mW_q$ and $\mK$ with $\mW_k$ respectively. The convolution is shared for $\mQ$ and $\mK$. The causal 1D structure of the convolution from the original mLSTM~\citep{beck2024xlstm} is necessary due to the causal 1D structure of language modeling. However, as images are neither causal nor 1D structures, we replace the causal 1D convolution with a 2D convolution (with kernel size 3). This allows the mLSTM to make better storage/retrieval decisions through the added local context.

\paragraph{(c) Positional Embedding} ViTs require positional embedding to tell the model where each patch is located in the image, suffering heavy performance losses if the position is not required~\citep{dosovitskiy2021vit,chu2023condPosEmbed}. The mLSTM is an autoregressive model, which makes it optional to add positional embeddings as it can recognize the position of the current patch based on how many patches have been processed. However, the ablation shows that it is nevertheless beneficial to provide this information explicitly as it improves segmentation results without hurting classification performance.

\paragraph{(d) Sequential vs.\ Parallel} Related architectures use a parallel design where a sequence is processed from multiple directions in a single block~\citep{zhou2024visionmamba,duan2024vrwkv}. We investigate a similar design where we apply both directions in parallel instead of sequentially. To keep parameters and FLOPS constant, we apply the directions akin to parallel transformer blocks~\citep{wang2021gptj} while halving the depth.

\begin{align}
\mY = \mX + \text{Block}_\theta(\mX) + \text{Flip}(\text{Block}_\phi(\text{Flip}(\mX)))
\end{align}

\subsection{Classification Design}

In order to perform classification from a sequence of tokens, it is common to aggregate information from the whole sequence, which is then used as input to a classification head.
The most common methods to do this aggregation are (i) adding a learnable [CLS] token to the input sequence or (ii) averaging all patch tokens to produce an [AVG] token. In ViTs, whether to use the [CLS] or [AVG] token is typically a hyperparameter, where both variants achieve comparable performances. On the contrary, other sequence models models often require specialized classification designs.
For example, Vim~\citep{zhou2024visionmamba} requires the [CLS] token to be in the middle of the sequence, suffering heavy performance losses if other classification designs, e.g., an [AVG] token or two [CLS] tokens at start and end of the sequence, are employed.

We explore different classification designs for ViL in Table~\ref{tab:pooling}.
(a) We choose concatenating the first and last patch as aggregation method due to its  strong classification performance. As our final models also perform well in semantic segmentation (see Table~\ref{tab:ade20k}), we do not retrain models with [AVG] aggregation even though the ablation suggests that this could boost performance even further for segmentation tasks. (b) Adding learnable [CLS] tokens show no benefit. Therefore, we do not use any [CLS] tokens for ViL.

\begin{table}[h]
\centering
\subfloat[
\textbf{Patch-based Aggregation}
]{
\centering
\begin{minipage}{0.45\linewidth}
\begin{center}
\begin{tabular}{lcc}
Aggregation & IN1K & ADE20K \\
\hline
Bilateral Mean   & 73.0 & 31.5 \\
\default{Bilateral Concat} & \default{\textbf{73.7}} & \default{31.7} \\
$[$AVG$]$        & 72.6 & \textbf{32.8} \\
Center [AVG]     & 72.4 & 32.1 \\
\end{tabular}
\end{center}
\end{minipage}
}
\subfloat[
\textbf{[CLS]-based Aggregation}
]{
\centering
\begin{minipage}{0.45\linewidth}
\begin{center}
\begin{tabular}{lccc}
Aggregation & IN1K \\
\hline
\default{Concat Bilateral Patches} & \default{\textbf{73.7}}\\
Mid [CLS] & 71.8 \\
Bilateral [CLS] & 73.5 \\
Mid + Bilateral [CLS] & 73.0 \\
\end{tabular}
\end{center}
\end{minipage}
}
\caption{ Classification design. (a) ViL aggregates classification information well in the first and the last patches (bilateral), leading to good classification performance if the first and last patches are averaged or concatenated. Averaging all patches ([AVG]) or the 4 center patches (Center [AVG]) results in strong segmentation performances but lackluster classification performances. (b) Adding learnable [CLS] tokens to the start and end of the input sequence (Bilateral [CLS]) offers no benefit over simply using the first and the last patch. Incorporating a [CLS] token in the middle of the sequence, akin to Vim~\citep{zhou2024visionmamba}, does not improve performance. \colorbox{defaultcolor}{Default settings}}
\label{tab:pooling}
\end{table}

\section{Limitations and Future Work}

The biggest limitation of ViL is the current lack of an optimized hardware implementation of the mLSTM, which results in longer runtimes than ViTs, which have multiple optimized hardware implementations~\citep{dao2022flashattention,dao2023flashattention2}. This makes a runtime/throughput analysis of models, a vital metric to judge practicability, difficult as the practical relevance of inefficient implementations is quite low. As a proxy, we report FLOP counts, where ViL is comparable to ViT on low-resolution tasks and far better than ViT on high-resolution tasks due to its linear complexity. While FLOPS are far from an optimal proxy for runtime/throughput, they suggest that ViL can be much faster than ViT on high-resolution tasks once an optimized hardware implementation exists. Note that ViL is already faster than Vim (see Appendix~\ref{app:runtime_vim_vs_vil}) despite its optimized hardware implementation.

This limitation snowballs in multiple other directions. For example, scaling model size further, tuning hyperparameters, training on larger datasets, exploring self-supervised pre-training or investigating hierarchical architectures are all interesting avenues for future work that are currently quite costly due to the lack of an optimized hardware implementation.

Please note that this is merely a technical limitation, not a methodical one as the mLSTM is heavily parallelizable. However, implementing fast compute kernels in CUDA~\citep{cuda} or Triton~\citep{tillet2019triton} is highly non-trivial as it requires expert hardware architecture knowledge, advanced implementation skills and potentially multiple development cycles to iron out numerical inaccuracies or instabilities.

However, the results of recent linear attention mechanisms show impressive FLOPS utilization (e.g., \cite{yang2024gla}). As the mLSTM can be parallelized with similar techniques it is only a matter of time that the mLSTM achieves a similar FLOPS utilization, which will make the mLSTM faster than transformers once an efficient hardware implementation is available.

Additionally, we made a significant effort to make our architecture as efficient as possible, using the tools that are currently available to us. Notably, our architecture is already much faster (up to 70\%) than Vim~\citep{zhou2024visionmamba} despite Vim using a custom CUDA kernel, as shown in Appendix~\ref{app:runtime_vim_vs_vil}. For reference, in language modeling, Mamba is roughly on-par with transformers in terms of speed and 4x faster than than the xLSTM (as mentioned in \cite{beck2024xlstm}), again, due to the current lack of efficient hardware implementation of the mLSTM. These considerations further underline the potential of our simple and efficient design for vision applications.

\section{Related Work}

\paragraph{Generic Vision Backbones.}
The inductive bias of CNNs~\citep{fukushima1980neocognitron,LeCun1998cnn} has demonstrated ground-breaking advancements in computer vision~\citep{krizhevsky2012alexnet} in the early deep learning days. Features of CNNs have been found to learn generic visual features that can be used for a variety of tasks~\citep{donahue2014transfer}. Subsequently, countless works improved various aspects such as architectures~\citep{szegedy2015inception,he2016resnet,huang2017densenet,tan2019efficientnet,liu2022convnext} or pre-training strategy~\citep{doersch2015ssl,noroozi2016jigsaw,zhang2016colorization,gidaris2018rotationpred,chen2020simclr,grill2020byol}.

\paragraph{Sequence Models in Vision.} The introduction of transformers~\citep{vaswani2017attention} demonstrated exceptional scalability in language processing, which motivated the vision community to explore transformers also in computer vision~\citep{chen2020imagegpt,cordonnier2020previt} but was applied on pixels or small patches which inhibited large costs due to the quadratic complexity of self-attention. This restriction was alleviated by the seminal work Vision Transformers (ViTs)~\citep{dosovitskiy2021vit} by using larger patches to aggregate local information and reduce training costs. Similar to CNNs, lots of work improved on the ViT architecture by refining training procedures~\citep{touvron2021deit,touvron2021cait,touvron2022deit3,dino2021caron,bao2021beit,xie2022simmim,he2022mae}. The recent advancement of autoregressive models in language processing~\citep{gu2023mamba,peng2023rwkv} has also gathered interest in the vision community~\citep{zhou2024visionmamba,duan2024vrwkv} due to the linear scaling property which allows applications to high-resolution tasks such as medical imaging~\citep{ma2024umamba_medical} or video understanding~\citep{li2024videomamba}.

\section{Conclusion}

Motivated by the success of xLSTM in language modeling, we introduced ViL, an adaption of the xLSTM architecture to vision tasks. ViL processes a sequence of patch tokens in alternating fashion. Odd blocks process image patches row-wise from top left to bottom right and even blocks go row-wise from bottom right to top left. Our new architecture outperforms SSM-based vision architectures, other autoregressive vision architectures and also optimized ViT models on ImageNet-1K classification, VTAB-1K transfer classification and ADE20K semantic segmentation. Remarkably, ViL is able to outperform ViT training pipelines, which are the result of years of hyperparameter tuning and transformer improvements.

In the future, we see potential in applying ViL when high-resolution images are needed for optimal performance, such as semantic segmentation or medical imaging. In these settings, transformers suffer from high computational costs due to the quadratic complexity of self-attention, where the linear complexity of ViL allows compute efficient processing of long sequences. Additionally, improving pre-training schemes (e.g., via self-supervised learning), exploring better hyperparameter settings or investigating hierarchical architectures are promising future directions.

\section*{Acknowledgments}

We acknowledge EuroHPC Joint Undertaking for awarding us access to Karolina at IT4Innovations, Czech Republic, MeluXina at LuxProvide, Luxembourg, Leonardo at CINECA, Italy and LUMI at CSC, Finland.

The ELLIS Unit Linz, the LIT AI Lab, the Institute for Machine Learning, are supported by the Federal State Upper Austria. We thank the projects Medical Cognitive Computing Center (MC3), INCONTROL-RL (FFG-881064), PRIMAL (FFG-873979), S3AI (FFG-872172), DL for GranularFlow (FFG-871302), EPILEPSIA (FFG-892171), AIRI FG 9-N (FWF-36284, FWF-36235), AI4GreenHeatingGrids (FFG- 899943), INTEGRATE (FFG-892418), ELISE (H2020-ICT-2019-3 ID: 951847), Stars4Waters (HORIZON-CL6-2021-CLIMATE-01-01). We thank Audi.JKU Deep Learning Center, TGW LOGISTICS GROUP GMBH, Silicon Austria Labs (SAL), FILL Gesellschaft mbH, Anyline GmbH, Google, ZF Friedrichshafen AG, Robert Bosch GmbH, UCB Biopharma SRL, Merck Healthcare KGaA, Verbund AG, GLS (Univ. Waterloo), Software Competence Center Hagenberg GmbH, Borealis AG, T\"{U}V Austria, Frauscher Sensonic, TRUMPF and the NVIDIA Corporation.

\clearpage

\bibliography{main}
\bibliographystyle{iclr2025_conference}
\newpage
\clearpage
\appendix

\section{Extended Results}

\subsection{Runtime Comparison of ViL vs Vim} \label{app:runtime_vim_vs_vil}

We compare the runtime to train ViL and Vim~\citep{zhou2024visionmamba} for 10 ImageNet-1K epochs in Table~\ref{tab:runtime_vim_vs_vil}. We follow the scaling procedure of ViTs, using 192 (T), 384 (S), 768 (B), 1024 (L) as hidden dimension where the (L)arge scale doubles the number of blocks.

\begin{table}[h]
\centering
\begin{tabular}{llcccc}
Model & \multicolumn{1}{l}{Optimization} & (T)iny & (S)mall & (B)ase & (L)arge \\
\hline
Vim~\citep{zhou2024visionmamba} & custom CUDA kernel & 7.3h & 14.0h & 28.2h & 76.4h \\
ViL & \texttt{torch.compile} & 5.0h & 8.7h & 16.6h & 45.1h \\
\hline
\multicolumn{2}{l}{Speedup of ViL compared to Vim} & 45\% & 61\% & 69\% & 69\% \\
\end{tabular}
\vspace{1em}
\caption{ Runtime comparisons between Vim~\citep{zhou2024visionmamba} and ViL. ViL is up to 69\% faster despite the current lack of a optimized hardware implementation. As mLSTM (and ViL) can be parallelized analogous to FlashAttention~\citep{dao2022flashattention,dao2023flashattention2} via custom hardware optimizations, ViL will become even faster in the future. Runtimes denote the training time for 10 ImageNet-1K epochs and are extrapolated from short benchmark runs on a single A100-80GB-PCIe using float16 precision and 224x224 images. }
\label{tab:runtime_vim_vs_vil}
\end{table}

\subsection{Impact of Longer Training}

We investigate the impact of training for a longer duration in Table~\ref{tab:longer_training}.

\begin{table}[h]
\centering
\begin{tabular}{lcccc}
Model & Epochs & IN-1K ACC & VTAB-1K & ADE20K mIoU \\
\hline
DeiT-III-T & 400 & 75.6 & 67.0 & 39.1 \\
DeiT-III-T & 800 & 76.2 & 67.1 & 39.8 \\
\hline
ViL-T & 400 & 77.2 & 67.8 & 40.9 \\
ViL-T & 800 & 78.3 & 68.3 & 41.2 \\
\end{tabular}
\vspace{1em}
\caption{Performance comparison of tiny models trained for 400 and 800 epochs. ADE20K mIoU uses single-scale evaluation. All settings follow the ones used in the main paper. }
\label{tab:longer_training}
\end{table}

\subsection{VTAB-1K Individual Dataset Results}

Table~\ref{tab:vtab1k_full} presents accuracies for each individual dataset of the VTAB-1K benchmark.

\begin{table}[h]
\centering
\resizebox{\textwidth}{!}{
\begin{tabular}{lccccccccccccccccccc}
\multicolumn{1}{c}{}&\multicolumn{7}{c}{{\textbf{Natural}}}&\multicolumn{4}{c}{{\textbf{Specialized}}}&\multicolumn{8}{c}{{\textbf{Structured}}}\\
\cmidrule(rl){2-8}\cmidrule(rl){9-12}\cmidrule(rl){13-20}
&\multicolumn{1}{c}{\rotatebox[origin=c]{90}{Cifar100}}
&\multicolumn{1}{c}{\rotatebox[origin=c]{90}{Caltech101}}
&\multicolumn{1}{c}{\rotatebox[origin=c]{90}{DTD}}
&\multicolumn{1}{c}{\rotatebox[origin=c]{90}{Flower102}}
&\multicolumn{1}{c}{\rotatebox[origin=c]{90}{Pets}}
&\multicolumn{1}{c}{\rotatebox[origin=c]{90}{SVHN}}
&\multicolumn{1}{c}{\rotatebox[origin=c]{90}{Sun397}}
&\multicolumn{1}{c}{\rotatebox[origin=c]{90}{Camelyon}}
&\multicolumn{1}{c}{\rotatebox[origin=c]{90}{EuroSAT}}
&\multicolumn{1}{c}{\rotatebox[origin=c]{90}{Resisc45}}
&\multicolumn{1}{c}{\rotatebox[origin=c]{90}{Retinopathy}}
&\multicolumn{1}{c}{\rotatebox[origin=c]{90}{Clevr-Count}}
&\multicolumn{1}{c}{\rotatebox[origin=c]{90}{Clevr-Dist}}
&\multicolumn{1}{c}{\rotatebox[origin=c]{90}{DMLab}}
&\multicolumn{1}{c}{\rotatebox[origin=c]{90}{KITTI-Dist}}
&\multicolumn{1}{c}{\rotatebox[origin=c]{90}{dSpr-Loc}}
&\multicolumn{1}{c}{\rotatebox[origin=c]{90}{dSpr-Ori}}
&\multicolumn{1}{c}{\rotatebox[origin=c]{90}{sNORB-Azim}}
&\multicolumn{1}{c}{\rotatebox[origin=c]{90}{sNORB-Ele}}\\
\hline
DeiT-T  & \cellcolor[rgb]{1.00,1.00,0.90}47.7 & \cellcolor[rgb]{1.00,1.00,0.90}86.4 & \cellcolor[rgb]{1.00,1.00,0.90}63.7 & \cellcolor[rgb]{1.00,1.00,0.90}85.6 & \cellcolor[rgb]{1.00,1.00,0.90}87.0 & \cellcolor[rgb]{0.91,0.96,0.84}78.4 & \cellcolor[rgb]{1.00,1.00,0.90}35.3 & \cellcolor[rgb]{1.00,1.00,0.90}83.0 & \cellcolor[rgb]{0.94,0.97,0.86}93.4 & \cellcolor[rgb]{1.00,1.00,0.90}80.9 & \cellcolor[rgb]{1.00,1.00,0.90}70.7 & \cellcolor[rgb]{1.00,1.00,0.90}71.7 & \cellcolor[rgb]{0.68,0.87,0.70}60.3 & \cellcolor[rgb]{1.00,1.00,0.90}43.1 & \cellcolor[rgb]{0.98,0.99,0.88}78.5 & \cellcolor[rgb]{0.23,0.69,0.43}\textbf{67.9} & \cellcolor[rgb]{1.00,1.00,0.90}41.6 & \cellcolor[rgb]{0.51,0.80,0.60}30.6 & \cellcolor[rgb]{0.82,0.93,0.79}32.7\\
DeiT-III-T  & \cellcolor[rgb]{1.00,1.00,0.90}52.3 & \cellcolor[rgb]{0.26,0.70,0.45}90.1 & \cellcolor[rgb]{1.00,1.00,0.90}62.7 & \cellcolor[rgb]{0.87,0.95,0.82}88.8 & \cellcolor[rgb]{1.00,1.00,0.90}87.5 & \cellcolor[rgb]{0.49,0.79,0.59}\textbf{83.7} & \cellcolor[rgb]{1.00,1.00,0.90}37.9 & \cellcolor[rgb]{0.97,0.99,0.88}83.2 & \cellcolor[rgb]{1.00,1.00,0.90}93.1 & \cellcolor[rgb]{0.99,0.99,0.89}81.1 & \cellcolor[rgb]{1.00,1.00,0.90}72.9 & \cellcolor[rgb]{0.75,0.90,0.75}76.6 & \cellcolor[rgb]{0.61,0.84,0.66}60.8 & \cellcolor[rgb]{0.92,0.97,0.85}44.9 & \cellcolor[rgb]{0.88,0.95,0.83}79.1 & \cellcolor[rgb]{0.26,0.70,0.45}67.5 & \cellcolor[rgb]{0.97,0.99,0.88}48.1 & \cellcolor[rgb]{0.45,0.78,0.56}31.0 & \cellcolor[rgb]{0.77,0.91,0.76}33.3\\
Vim-T  & \cellcolor[rgb]{1.00,1.00,0.90}46.7 & \cellcolor[rgb]{1.00,1.00,0.90}86.3 & \cellcolor[rgb]{1.00,1.00,0.90}60.7 & \cellcolor[rgb]{1.00,1.00,0.90}84.0 & \cellcolor[rgb]{1.00,1.00,0.90}88.8 & \cellcolor[rgb]{1.00,1.00,0.90}76.1 & \cellcolor[rgb]{1.00,1.00,0.90}33.7 & \cellcolor[rgb]{1.00,1.00,0.90}82.2 & \cellcolor[rgb]{1.00,1.00,0.90}92.9 & \cellcolor[rgb]{1.00,1.00,0.90}75.2 & \cellcolor[rgb]{1.00,1.00,0.90}72.6 & \cellcolor[rgb]{1.00,1.00,0.90}59.8 & \cellcolor[rgb]{1.00,1.00,0.90}49.9 & \cellcolor[rgb]{1.00,1.00,0.90}39.3 & \cellcolor[rgb]{1.00,1.00,0.90}78.2 & \cellcolor[rgb]{1.00,1.00,0.90}51.2 & \cellcolor[rgb]{1.00,1.00,0.90}43.9 & \cellcolor[rgb]{1.00,1.00,0.90}26.9 & \cellcolor[rgb]{1.00,1.00,0.90}27.2\\
ViL-T  & \cellcolor[rgb]{0.90,0.96,0.84}\textbf{54.2} & \cellcolor[rgb]{0.23,0.69,0.43}\textbf{90.2} & \cellcolor[rgb]{0.62,0.85,0.67}\textbf{67.4} & \cellcolor[rgb]{0.62,0.85,0.67}\textbf{90.7} & \cellcolor[rgb]{0.89,0.96,0.83}\textbf{89.9} & \cellcolor[rgb]{0.66,0.86,0.69}81.6 & \cellcolor[rgb]{0.79,0.91,0.77}\textbf{41.1} & \cellcolor[rgb]{0.91,0.97,0.85}\textbf{83.4} & \cellcolor[rgb]{0.60,0.84,0.66}\textbf{94.2} & \cellcolor[rgb]{0.75,0.90,0.75}\textbf{82.7} & \cellcolor[rgb]{0.94,0.98,0.86}\textbf{73.1} & \cellcolor[rgb]{0.46,0.78,0.57}\textbf{80.7} & \cellcolor[rgb]{0.46,0.78,0.57}\textbf{61.8} & \cellcolor[rgb]{0.52,0.81,0.61}\textbf{49.4} & \cellcolor[rgb]{0.54,0.81,0.62}\textbf{81.3} & \cellcolor[rgb]{1.00,1.00,0.90}57.8 & \cellcolor[rgb]{0.68,0.87,0.70}\textbf{51.8} & \cellcolor[rgb]{0.40,0.75,0.53}\textbf{31.4} & \cellcolor[rgb]{0.65,0.86,0.68}\textbf{34.8}\\
\hline
DeiT-S  & \cellcolor[rgb]{0.71,0.88,0.72}57.0 & \cellcolor[rgb]{0.66,0.86,0.69}88.9 & \cellcolor[rgb]{0.54,0.81,0.62}68.2 & \cellcolor[rgb]{0.59,0.84,0.65}90.9 & \cellcolor[rgb]{0.71,0.88,0.72}90.8 & \cellcolor[rgb]{1.00,1.00,0.90}75.4 & \cellcolor[rgb]{0.69,0.88,0.71}42.1 & \cellcolor[rgb]{0.94,0.98,0.86}83.3 & \cellcolor[rgb]{0.69,0.87,0.71}\textbf{94.0} & \cellcolor[rgb]{0.59,0.83,0.65}83.8 & \cellcolor[rgb]{0.45,0.78,0.56}\textbf{74.0} & \cellcolor[rgb]{0.90,0.96,0.84}74.6 & \cellcolor[rgb]{0.97,0.99,0.88}58.3 & \cellcolor[rgb]{0.86,0.94,0.81}45.6 & \cellcolor[rgb]{1.00,1.00,0.90}78.2 & \cellcolor[rgb]{0.70,0.88,0.71}\textbf{61.9} & \cellcolor[rgb]{0.99,1.00,0.89}47.9 & \cellcolor[rgb]{1.00,1.00,0.90}27.1 & \cellcolor[rgb]{0.89,0.95,0.83}31.9\\
DeiT-III-S  & \cellcolor[rgb]{0.59,0.83,0.65}58.8 & \cellcolor[rgb]{0.76,0.90,0.75}88.6 & \cellcolor[rgb]{0.61,0.84,0.66}67.5 & \cellcolor[rgb]{0.59,0.84,0.65}90.9 & \cellcolor[rgb]{0.53,0.81,0.61}\textbf{91.7} & \cellcolor[rgb]{0.43,0.77,0.55}\textbf{84.4} & \cellcolor[rgb]{0.58,0.83,0.64}43.3 & \cellcolor[rgb]{0.63,0.85,0.67}84.4 & \cellcolor[rgb]{1.00,1.00,0.90}92.6 & \cellcolor[rgb]{0.78,0.91,0.76}82.5 & \cellcolor[rgb]{0.72,0.89,0.73}73.5 & \cellcolor[rgb]{0.76,0.90,0.75}76.5 & \cellcolor[rgb]{1.00,1.00,0.90}57.9 & \cellcolor[rgb]{0.80,0.92,0.78}46.2 & \cellcolor[rgb]{0.91,0.97,0.85}78.9 & \cellcolor[rgb]{0.98,0.99,0.89}58.3 & \cellcolor[rgb]{0.85,0.94,0.80}49.7 & \cellcolor[rgb]{1.00,1.00,0.90}23.7 & \cellcolor[rgb]{1.00,1.00,0.90}27.5\\
Vim-S  & \cellcolor[rgb]{0.98,0.99,0.88}53.0 & \cellcolor[rgb]{1.00,1.00,0.90}87.2 & \cellcolor[rgb]{0.98,0.99,0.88}64.1 & \cellcolor[rgb]{1.00,1.00,0.90}86.8 & \cellcolor[rgb]{0.81,0.92,0.78}90.3 & \cellcolor[rgb]{1.00,1.00,0.90}65.8 & \cellcolor[rgb]{0.92,0.97,0.85}39.7 & \cellcolor[rgb]{1.00,1.00,0.90}82.4 & \cellcolor[rgb]{0.94,0.97,0.86}93.4 & \cellcolor[rgb]{1.00,1.00,0.90}78.0 & \cellcolor[rgb]{0.94,0.98,0.86}73.1 & \cellcolor[rgb]{1.00,1.00,0.90}63.1 & \cellcolor[rgb]{1.00,1.00,0.90}53.2 & \cellcolor[rgb]{1.00,1.00,0.90}42.3 & \cellcolor[rgb]{1.00,1.00,0.90}78.2 & \cellcolor[rgb]{1.00,1.00,0.90}54.1 & \cellcolor[rgb]{1.00,1.00,0.90}47.6 & \cellcolor[rgb]{1.00,1.00,0.90}27.1 & \cellcolor[rgb]{1.00,1.00,0.90}29.3\\
ViL-S  & \cellcolor[rgb]{0.42,0.76,0.54}\textbf{61.4} & \cellcolor[rgb]{0.43,0.77,0.55}\textbf{89.6} & \cellcolor[rgb]{0.43,0.77,0.55}\textbf{69.2} & \cellcolor[rgb]{0.35,0.73,0.50}\textbf{92.8} & \cellcolor[rgb]{0.53,0.81,0.61}\textbf{91.7} & \cellcolor[rgb]{0.89,0.95,0.83}78.7 & \cellcolor[rgb]{0.53,0.81,0.61}\textbf{43.8} & \cellcolor[rgb]{0.31,0.72,0.48}\textbf{85.5} & \cellcolor[rgb]{0.73,0.89,0.73}93.9 & \cellcolor[rgb]{0.50,0.80,0.60}\textbf{84.4} & \cellcolor[rgb]{0.72,0.89,0.73}73.5 & \cellcolor[rgb]{0.23,0.69,0.43}\textbf{84.0} & \cellcolor[rgb]{0.23,0.69,0.43}\textbf{63.4} & \cellcolor[rgb]{0.35,0.74,0.50}\textbf{51.3} & \cellcolor[rgb]{0.23,0.69,0.43}\textbf{83.3} & \cellcolor[rgb]{0.77,0.91,0.76}61.0 & \cellcolor[rgb]{0.40,0.76,0.53}\textbf{55.4} & \cellcolor[rgb]{0.26,0.70,0.44}\textbf{32.4} & \cellcolor[rgb]{0.59,0.83,0.65}\textbf{35.5}\\
\hline
DeiT-B  & \cellcolor[rgb]{0.39,0.75,0.53}61.8 & \cellcolor[rgb]{0.36,0.74,0.51}89.8 & \cellcolor[rgb]{0.61,0.84,0.66}67.5 & \cellcolor[rgb]{0.23,0.69,0.43}\textbf{93.7} & \cellcolor[rgb]{0.35,0.74,0.50}92.6 & \cellcolor[rgb]{0.43,0.77,0.55}84.4 & \cellcolor[rgb]{0.37,0.74,0.51}45.6 & \cellcolor[rgb]{0.37,0.74,0.51}85.3 & \cellcolor[rgb]{0.23,0.69,0.43}\textbf{95.1} & \cellcolor[rgb]{0.23,0.69,0.43}\textbf{86.3} & \cellcolor[rgb]{0.34,0.73,0.49}74.2 & \cellcolor[rgb]{0.68,0.87,0.70}77.7 & \cellcolor[rgb]{0.74,0.89,0.74}59.9 & \cellcolor[rgb]{0.72,0.88,0.72}47.2 & \cellcolor[rgb]{0.48,0.79,0.58}\textbf{81.7} & \cellcolor[rgb]{0.71,0.88,0.72}61.7 & \cellcolor[rgb]{0.71,0.88,0.72}51.4 & \cellcolor[rgb]{0.59,0.83,0.65}30.0 & \cellcolor[rgb]{0.53,0.81,0.61}36.2\\
DeiT-III-B  & \cellcolor[rgb]{0.32,0.72,0.48}62.9 & \cellcolor[rgb]{0.43,0.77,0.55}89.6 & \cellcolor[rgb]{0.39,0.75,0.53}69.6 & \cellcolor[rgb]{0.23,0.69,0.43}\textbf{93.7} & \cellcolor[rgb]{0.23,0.69,0.43}\textbf{93.2} & \cellcolor[rgb]{0.23,0.69,0.43}\textbf{87.0} & \cellcolor[rgb]{0.23,0.69,0.43}\textbf{47.1} & \cellcolor[rgb]{0.23,0.69,0.43}\textbf{85.8} & \cellcolor[rgb]{0.65,0.86,0.68}94.1 & \cellcolor[rgb]{0.33,0.73,0.49}85.6 & \cellcolor[rgb]{0.61,0.84,0.66}73.7 & \cellcolor[rgb]{0.48,0.79,0.58}80.5 & \cellcolor[rgb]{0.52,0.80,0.61}61.4 & \cellcolor[rgb]{0.61,0.84,0.66}48.4 & \cellcolor[rgb]{0.60,0.84,0.66}80.9 & \cellcolor[rgb]{0.50,0.80,0.59}\textbf{64.4} & \cellcolor[rgb]{0.42,0.77,0.55}55.1 & \cellcolor[rgb]{0.56,0.82,0.63}30.2 & \cellcolor[rgb]{0.90,0.96,0.84}31.8\\
ViL-B  & \cellcolor[rgb]{0.23,0.69,0.43}\textbf{64.3} & \cellcolor[rgb]{0.29,0.71,0.47}\textbf{90.0} & \cellcolor[rgb]{0.23,0.69,0.43}\textbf{71.1} & \cellcolor[rgb]{0.27,0.70,0.45}93.4 & \cellcolor[rgb]{0.59,0.83,0.65}91.4 & \cellcolor[rgb]{0.81,0.92,0.78}79.6 & \cellcolor[rgb]{0.27,0.71,0.46}46.6 & \cellcolor[rgb]{0.54,0.81,0.62}84.7 & \cellcolor[rgb]{0.56,0.82,0.63}94.3 & \cellcolor[rgb]{0.36,0.74,0.51}85.4 & \cellcolor[rgb]{0.23,0.69,0.43}\textbf{74.4} & \cellcolor[rgb]{0.25,0.69,0.44}\textbf{83.7} & \cellcolor[rgb]{0.42,0.76,0.54}\textbf{62.1} & \cellcolor[rgb]{0.23,0.69,0.43}\textbf{52.7} & \cellcolor[rgb]{0.59,0.83,0.65}81.0 & \cellcolor[rgb]{0.60,0.84,0.66}63.1 & \cellcolor[rgb]{0.23,0.69,0.43}\textbf{57.6} & \cellcolor[rgb]{0.23,0.69,0.43}\textbf{32.6} & \cellcolor[rgb]{0.23,0.69,0.43}\textbf{39.9}\\
\end{tabular}}
\vspace{1em}
\caption{Results on all datasets of the VTAB-1K~\citep{zhai2019vtab} benchmark.}
\label{tab:vtab1k_full}
\end{table}

\clearpage
\subsection{Robustness and Domain Generalization} \label{sec:robustness}

Table~\ref{tab:robustness} presents robustness and OOD evaluations of ImageNet-1K pre-trained classifiers.

\begin{table}[h]
\centering
\begin{tabular}{lccccc}
Model & IN-C ($\downarrow$) & IN-A ($\uparrow$) & IN-R ($\uparrow$) & Sketch ($\uparrow$) & Validation ($\uparrow$) \\
\hline
DeiT-T & 69.7 & 7.6 & 32.7 & 19.9 & 72.2 \\
DeiT-III-T & 65.0 & 11.7 & 39.4 & 27.4 & 76.2 \\
Vim-T & 61.8 & 9.6 & 38.8 & 26.9 & 76.1 \\
ViL-T & \textbf{59.6} & \textbf{15.2} & \textbf{42.2} & \textbf{30.0} & \textbf{78.3} \\
\hline
DeiT-S & 54.4 & 19.6 & 41.9 & 29.1 & 79.8 \\
DeiT-III-S & \textbf{50.1} & 23.2 & 46.6 & \textbf{35.4} & 81.4 \\
Vim-S & 51.5 & 19.7 & 44.8 & 32.5 & 80.5 \\
ViL-S & 50.6 & \textbf{23.8} & \textbf{47.9} & 35.2 & \textbf{81.5} \\
\hline
DeiT-B & 48.6 & 27.9 & 44.6 & 32.0 & 81.8 \\
DeiT-III-B & \textbf{42.7} & \textbf{36.5} & \textbf{54.1} & \textbf{41.1} & \textbf{83.8} \\
ViL-B & {45.3} & {30.9} & {51.9} & {39.0} & {82.4} \\
\end{tabular}
\vspace{1em}
\caption{Robustness and OOD evaluations on ImageNet-C(orruption)~\citep{hendrycks19imagenetc}, ImageNet-A(dversarial)~\citep{hendrycks2021imageneta}, ImageNet-R(endition)~\citep{hendrycks2021imagenetr} and ImageNet-Sketch~\citep{wang2019imagenetsketch}.. For ImageNet-C, we report the mean corruption error~\citep{hendrycks19imagenetc} with AlexNet~\citep{krizhevsky2012alexnet} as baseline.}
\label{tab:robustness}
\end{table}

\section{Implementation Details} \label{sec:implementation_details}

\subsection{Hardware} \label{appendix_hardware}
We train models on servers with either 8xA100 or 4xA100 nodes.

We estimate the total number of A100 GPU-hours used for this project to be 38K hours. This estimate includes initial exploration, method development, analysis and evaluations.

\subsection{FLOPS Calculation} \label{sec:flops}

We use the \texttt{fvcore}\footnote{https://github.com/facebookresearch/fvcore} library to count FLOPS and report FLOPS of the mLSTM chunkwise form as described in Section~\ref{sec:method_vil}. 
For the parallel parts, we report FLOPS for a complexity of $\mathcal{O}\big((\frac{S}{2}+1)Sd\big)$ because the upper triangular entries of the $\mathbf{QK}$ matrix do not need to be calculated due to the causal structure. We justify this by the fact that FlashAttention-2~\citep{dao2023flashattention2} is approximately 1.7x faster with a causal mask than without. Therefore, an optimized hardware implementation of the mLSTM could also omit the calculation of the upper triangular part of $\mathbf{QK}$.

As Vim~\citep{zhou2024visionmamba} does not report FLOPS and their model makes use of CUDA kernels (which are not counted as FLOPS by \texttt{fvcore}), we replace all calls to CUDA kernels with their reference PyTorch implementation and count the FLOPS with \texttt{fvcore}.

For the total pre-training compute in Figure~\ref{fig:flops_over_performance}, we consider an efficient implementation of stochastic depth~\citep{huang16stochasticdepth,touvron2023efficient_stochastic_depth} which omits the calculation of a dropped block instead of masking it. Therefore, we change the implementation of ViT~\citep{dosovitskiy2021vit}
to use our efficient stochastic depth implementation. Vim does not use stochastic depth for training as they only train tiny and small models.

\clearpage
\subsection{ViL Hyperparameters} \label{sec:vil_hyperparameters}
Table~\ref{tab:hyperparams_ViL_pretrain} shows detailed hyperparameters used to train ViL models.

\begin{table}[h]
\begin{center}
\begin{tabular}{l|c}
Parameter & Value  \\
\hline
Epochs & 800 (T), 400 (S/B) $\rightarrow$ 20 (T, S), 5 (B) \\
Batch size & 2048 $\rightarrow$ 1024 \\
Model & \\
\quad Patch size & 16x16 \\
\quad Latent dimension & 192 (T), 384 (S), 768 (B) \\
\quad Depth & 24 \\
\quad Pooling & Bilateral Concat \\
Stochastic depth &  \\
\quad Peak rate & 0 (T), 0.05 (S), 0.2 (B) \\
\quad Layer-wise Decay & \xmark \\
Optimizer & AdamW \\
\quad Base Learning rate & 1e-3 $\rightarrow$ 1e-5 \\
\quad Linear LR Scaling Divisor & 1024 \\
\quad Weight decay & 0.05\\
\quad Momentum & $\beta_1=0.9,\beta_2=0.999$ \\
\quad Gradient Norm Clip & 1.0 \\
Precision & mixed \texttt{bfloat16} \\
\quad Backend & \texttt{torch.autocast} \\
Learning rate schedule & cosine decay \\
\quad Warmup schedule & linear \\
\quad Warmup epochs & 5 $\rightarrow$ 5 (T, S), 1 (B) \\
\quad End LR & 1e-6 \\
Label smoothing & \xmark \\
Train Data Augmentation & \\
\quad {RandomResizedCrop} & 192 $\rightarrow$ 224 \\
\qquad {Scale} & [0.08, 1.0] \\
\qquad {Interpolation} & bicubic \\
\quad {RandomHorizontalFlip} & $p=0.5$ \\
\quad {3-Augment} &  \\
\qquad {Gaussian Blur $\sigma$} & [0.1, 2.0] \\
\qquad {ColorJitter} & [0.3, 0.3, 0.3, 0.0] \\
\quad {Normalize} & ImageNet-1K statistics \\
\quad {Mixup} $\alpha$ & 0.8 \\
\quad {Cutmix} $\alpha$ & 1.0 \\
Test Data Augmentation & \\
\quad {Resize} & 192 $\rightarrow$ 224 \\
\qquad {Interpolation} & bicubic \\
\quad {CenterCrop} & 192 $\rightarrow$ 224 \\
\quad {Normalize} & ImageNet-1K statistics \\
\end{tabular}
\end{center}
\caption{ Hyperparameters for training ViL on ImageNet-1K, inspired by DeiT-III~\citep{touvron2022deit3}. We follow the best setting from DeiT-III~\citep{touvron2022deit3} and pre-train on 192 resolution followed by a short fine-tuning on 224 resolution (indicated by $\rightarrow$).}
\label{tab:hyperparams_ViL_pretrain}
\end{table}

\clearpage
\subsection{Fine-tuning on VTAB-1K}

For fine-tuning models on VTAB-1K we provide the hyperparameters in Table~\ref{tab:hyperparams_finetune_vtab1k}.
We search for the best learning rate for each dataset by fine-tuning the model 25 times (5 learning rates with 5 seeds each) on the 800 training samples and evaluating them on the 200 validation samples. With the best learning rate, we then train each model 5 times on concatenation of training and validation split, evaluate on the test split and report the average accuracy.

\begin{table}[h]
\begin{center}
\begin{tabular}{l|c}
Parameter & Value  \\
\hline
Epochs & 50 \\
Batch size & 64 \\
Seeds & 5 \\
Optimizer & AdamW \\
\quad Learning rate & [1e-3, 7.5e-4, 5.0e-4, 2.5e-4, 1.0e-4] \\
\quad Layer-wise lr deca & 0.65* \\
\quad Weight decay & 0.05\\
\quad Momentum & $\beta_1=0.9,\beta_2=0.999$ \\
Learning rate schedule & linear warmup $\rightarrow$ cosine decay \\
\quad Warmup epochs & 5 \\
Precision & mixed \texttt{bfloat16} \\
\quad Backend & \texttt{torch.autocast} \\
Data Augmentation & \\
\quad \tt{Resize} &  \\
\qquad \tt{interpolation} & bicubic \\
\qquad \tt{size} & 224x224 \\
\quad \tt{Normalize} & ImageNet-1K statistics \\
\end{tabular}
\end{center}
\caption{ Hyperparameters for fine-tuning on VTAB-1K. *For Vim and ViL we group two consecutive blocks for the layer-wise lr decay similar to how ViT considers a pair of attention and MLP block as a single ``layer'' for the decay. }
\label{tab:hyperparams_finetune_vtab1k}
\end{table}

\clearpage
\subsection{ADE20K Semantic Segmentation Fine-tuning}

We fine-tune models on ADE20K~\citep{zhou2019ade20k} using an UperNet~\citep{xiao2018upernet} head. We follow common practices and fine-tune on 512x512 resolution, where we interpolate the absolute positional embedding from 224x224 to 512x512. For ViTs, we add relative position biases to the attention layers (initialized to 0)~\citep{he2022mae}. Table~\ref{tab:hyperparams_finetune_ade20k} lists detailed hyperparameters.

\begin{table}[h]
\begin{center}
\begin{tabular}{l|c}
Parameter & Value  \\
\hline
Updates & 160K \\
Batch size & 16 \\
UperNet & \\
\quad Auxiliary & \\
\qquad Weight & 0.4 \\
\qquad Input Block & 8* \\
\qquad Dimension & 192 (T), 384 (S, B) \\
\quad Decoder & \\
\qquad Weight & 1.0 \\
\qquad Input Blocks & [4, 6, 8, 12]* \\
\qquad Dimension & 192 (T), 384 (S, B) \\
Stochastic depth &  \\
\quad Peak rate & 0 (T), 0.05 (S), 0.1 (B) \\
\quad Layer-wise Decay & \cmark \\
Optimizer & AdamW \\
\quad Learning rate & 5e-4 \\
\quad Linear LR Scaling Divisor & 16 \\
\quad Layer-wise lr decay & 0.65* \\
\quad Weight decay & 0.05\\
\quad Momentum & $\beta_1=0.9,\beta_2=0.999$ \\
Learning rate schedule & linear warmup $\rightarrow$ cosine decay \\
\quad Warmup updates & 1500 \\
Precision & mixed \texttt{float16} \\
\quad Backend & \texttt{torch.autocast} \\
Train Data Augmentation & \\
\quad \tt{RandomResize} &  \\
\qquad \tt{interpolation} & bicubic \\
\quad \tt{RandomCrop} &  \\
\qquad \tt{size} & 512x512 \\
\quad \tt{RandomHorizontalFlip} &  \\
\quad \tt{ColorJitter} & 0.5 \\
\qquad \tt{brightness} & 0.5 \\
\qquad \tt{contrast} & 0.5 \\
\qquad \tt{saturation} & 0.5 \\
\qquad \tt{hue} & 0.25 \\
\quad \tt{Normalize} & ImageNet-1K statistics \\
Evaluation & \\
\quad \tt{Stride} & 341 \\
\quad \tt{Multi-scale} &  \\
\qquad \tt{scale factors} & [0.75, 1.0, 1.25, 1.5, 1.75] \\
\qquad \tt{flip} & [True, False] \\
\end{tabular}
\end{center}
\caption{ Hyperparameters for fine-tuning on VTAB-1K. *For ViL we group two consecutive blocks into one similar to how a ViT block consists of a pair of attention and MLP block. }
\label{tab:hyperparams_finetune_ade20k}
\end{table}

\clearpage
\subsection{DeiT-III Reimplementation Hyperparameters}
Table~\ref{tab:hyperparams_ViL_pretrain} shows detailed hyperparameters used to train DeiT-III-T (reimpl.) from Table~\ref{tab:imagenet1k}. Our reimplementation easily outperforms older baselines like DeiT-II-T (+2.7\% ImageNet-1K accuracy) and is approximately even with the original on ADE20K (40.1 vs 39.8 on mIoU single-scale, 41.8 vs 42.2 mIoU multi-scale).

\begin{table}[h]
\begin{center}
\begin{tabular}{l|c}
Parameter & Value  \\
\hline
Epochs & 800 $\rightarrow$ 20 \\
Batch size & 2048 $\rightarrow$ 1024 \\
Model & \\
\quad Patch size & 16x16 \\
\quad Latent dimension & 192 \\
\quad Depth & 12 \\
\quad Pooling & [CLS] \\
Stochastic depth & \xmark \\
Layerscale & 1e-4 \\
Optimizer & AdamW \\
\quad Base Learning rate & 1e-3 $\rightarrow$ 1e-5 \\
\quad Linear LR Scaling Divisor & 1024 \\
\quad Weight decay & 0.05\\
\quad Momentum & $\beta_1=0.9,\beta_2=0.999$ \\
\quad Gradient Norm Clip & \xmark \\
Precision & mixed \texttt{bfloat16} \\
\quad Backend & \texttt{torch.autocast} \\
Learning rate schedule & cosine decay \\
\quad Warmup schedule & linear  \\
\quad Warmup epochs & 5 \\
\quad End LR & 1e-6 \\
Label smoothing & \xmark \\
Train Data Augmentation & \\
\quad {RandomResizedCrop} & 192 $\rightarrow$ 224 \\
\qquad {Scale} & [0.08, 1.0] \\
\qquad {Interpolation} & bicubic \\
\quad {RandomHorizontalFlip} & $p=0.5$ \\
\quad {3-Augment} &  \\
\qquad {Gaussian Blur $\sigma$} & [0.1, 2.0] \\
\qquad {ColorJitter} & [0.3, 0.3, 0.3, 0.0] \\
\quad {Normalize} & ImageNet-1K statistics \\
\quad {Mixup} $\alpha$ & 0.8 \\
\quad {Cutmix} $\alpha$ & 1.0 \\
Test Data Augmentation & \\
\quad {Resize} & 192 $\rightarrow$ 224 \\
\qquad {Interpolation} & bicubic \\
\quad {CenterCrop} & 192 $\rightarrow$ 224 \\
\quad {Normalize} & ImageNet-1K statistics \\
\end{tabular}
\end{center}
\caption{ Hyperparameters for training our reimplementation of DeiT-III-T~\citep{touvron2022deit3} on ImageNet-1K. The most significant change is that we reduce the learning rate from 3e-3 to 1e-3 as we found this to greatly improve performance. We make minor changes to the protocol such as using AdamW or no gradient clipping as models were stable without it. We follow the best setting from DeiT-III~\citep{touvron2022deit3} and pre-train on 192 resolution followed by a short fine-tuning on 224 resolution (indicated by $\rightarrow$). }
\label{tab:hyperparams_vittiny}
\end{table}

\end{document}